%% file: main.tex
\documentclass[runningheads]{llncs}

\usepackage{graphicx}
\usepackage{amsmath,bm,amssymb} 
\usepackage{url}
\usepackage{float}
\usepackage{epsfig} 
\usepackage{hyperref}
\usepackage{subcaption}
\usepackage{multirow}
\usepackage{booktabs}
\usepackage{xcolor}
\usepackage{bibnames}
\usepackage[symbol]{footmisc}

\begin{document}
\title{ComBiNet: Compact Convolutional Bayesian Neural Network for Image Segmentation}
\titlerunning{Compact Convolutional Bayesian Neural Network for Image Segmentation}
%
\author{Martin Ferianc\inst{1} \and
Divyansh Manocha\inst{2} \and
Hongxiang Fan\inst{3}\and Miguel Rodrigues\inst{1}}
\institute{University College London, London WC1E 7JE, U.K., \\ \email{\{martin.ferianc.19,m.rodrigues\}@ucl.ac.uk} \and
\email{divyanshmanocha@gmail.com} \and
Imperial College London, London SW7 2AZ, U.K., \email{h.fan17@imperial.ac.uk}}
\maketitle              
\begin{abstract}

Fully convolutional U-shaped neural networks have largely been the dominant approach for pixel-wise image segmentation. In this work, we tackle two defects that hinder their deployment in real-world applications: \textit{1)} Predictions lack uncertainty quantification that may be crucial to many decision-making systems; \textit{2)} Large memory storage and computational consumption demanding extensive hardware resources. To address these issues and improve their practicality we demonstrate a few-parameter compact Bayesian convolutional architecture, that achieves a marginal improvement in accuracy in comparison to related work using significantly fewer parameters and compute operations. The architecture combines parameter-efficient operations such as separable convolutions, bilinear interpolation, multi-scale feature propagation and Bayesian inference for per-pixel uncertainty quantification through Monte Carlo Dropout. The best performing configurations required fewer than 2.5 million parameters on diverse challenging datasets with few observations.

\end{abstract}

\input{introduction}
\input{related_work}
\input{method}
\input{experiments}
\input{conclusion}

\subsubsection*{Acknowledgements.}
We thank the ICANN 2021 reviewers for useful feedback. Martin Ferianc was sponsored through a scholarship from ICCS at UCL.

\bibliographystyle{splncs04}
\bibliography{main}
\end{document}

%% file: introduction.tex
\section{Introduction}\label{sec:introduction}

Image segmentation is the pixel-level computer vision task of segregating an image into discrete regions semantically. Among various algorithms, convolutional neural networks (CNNs) have been key to this task, demonstrating outstanding performance~\cite{huang2017densely,he2016deep,ronneberger2015u,jegou2017one,kendall2015bayesian,howard2017mobilenets,zhao2017pyramid,long2014fully,badrinarayanan2017segnet}. CNNs are able to express predictions as pixel-wise output masks by learning appropriate feature representations in an end-to-end fashion, while allowing processing inputs with various size. This is especially useful in inferring object support relationships for robotics, autonomous driving or healthcare, as well as scene geometry~\cite{kendall2015bayesian,mcallister2017concrete,ruiz2018survey}.

 \begin{figure}[t]
 \centering
 \includegraphics[width=1\linewidth]{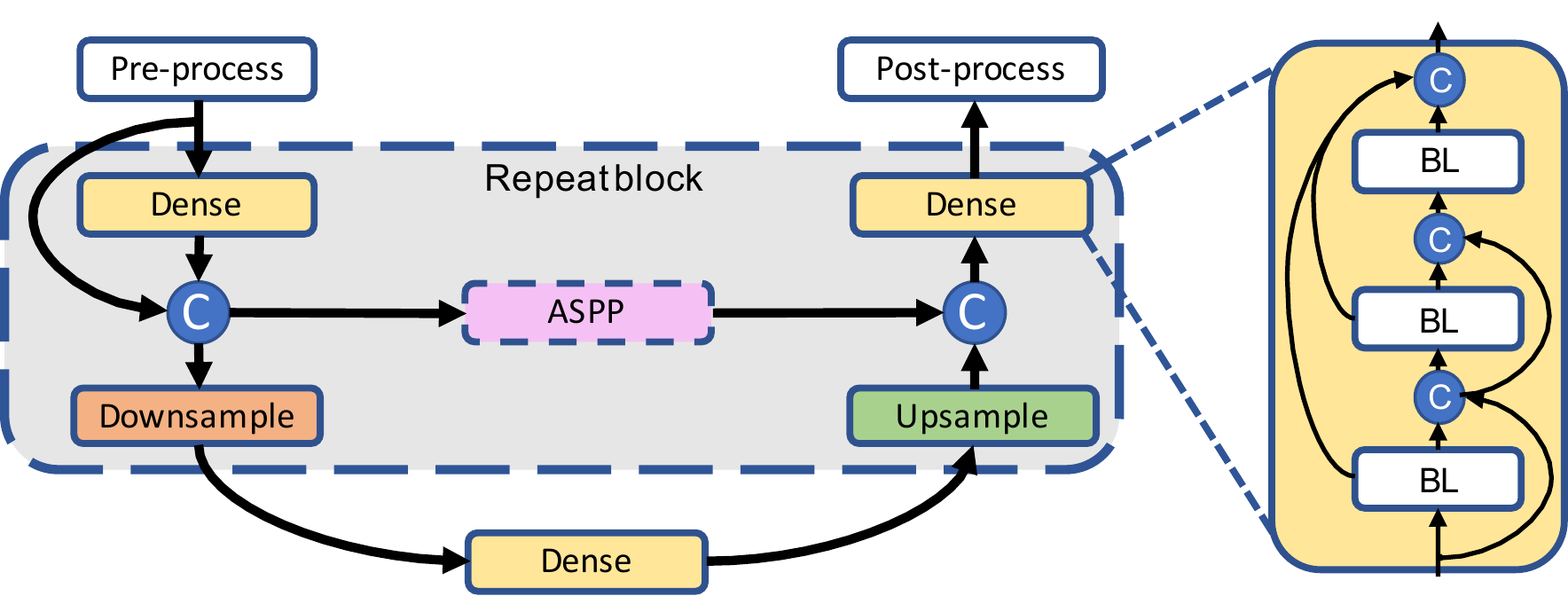}
     \caption{\textit{ComBiNet} an U-Net~\cite{ronneberger2015u,jegou2017one} like architecture consisting of Repeat blocks with different input scales and dilation rates in an Atrous Spatial Pyramid Pooling (ASPP) module. The block contains Dense feature extracting blocks, Downsampling to reduce the spatial dimensionality by $2\times$ and Upsampling for restoring it back after processing the features from a lower dimensionality. The context is transferred through an optional ASPP module and concatenated (C), pairing the same spatial resolution. On the right is the detail of the Dense block consisting of Basic Layers (BLs). The arrows represent data flow.}
     \label{fig:architecture}
 \end{figure}

A practical drawback of regular CNNs is that they are unable to capture their uncertainty which is crucial for many safety-critical applications~\cite{gal2016dropout}. Bayesian CNNs~\cite{gal2015bayesian} adopt Bayesian inference to provide a principled uncertainty estimation on top of the segmentation masks. However, as the research community seeks to improve accuracy and better capture information in a wider range of applications, potential CNN architectures become deeper and further connection-wise complicated~\cite{he2016deep,jegou2017one,liu2019auto,zhao2017pyramid}. As a result they are increasingly more compute and memory demanding and a regular modern CNN architecture cannot be easily adopted for Bayesian inference. As an analytical prediction of uncertainty is not tractable with such architectures, it is required to approximate it through Monte Carlo sampling with multiple runs through the network. The increased runtime cost, primarily due to sampling, has been a limiting factor of Bayesian CNNs in real-world image segmentation.

To address the aforementioned issues of lacking uncertainty quantification in regular CNNs and extensive execution cost, the contribution of this work is in improving the hardware performance of two-dimensional (2D) Bayesian CNNs for image segmentation, while also considering an efficient pixel-wise uncertainty quantification. Our approach builds on recent successes to improve software-hardware performance~\cite{he2016deep,jegou2017one,chen2018encoder,ronneberger2015u,howard2017mobilenets,kendall2015bayesian,zhang2019making} and extends these into a novel 2D Bayesian CNN architectural template as shown in Figure~\ref{fig:architecture}. Specifically, we focus on few-parameter/few-operation models which decrease the runtime cost of each feedforward pass, and present a compact design named \textit{ComBiNet}. Monte Carlo Dropout~\cite{gal2016dropout} is used for Bayesian inference. The novelty of our work is in demonstrating that it is possible to develop compressed models for 2D image segmentation while preserving uncertainty estimation capabilities, without compromising accuracy. We demonstrate ComBiNet's fine performance on the few-samples video-based CamVid~\cite{BrostowSFC:ECCV08} dataset and a database of darkfield microscopy images~\cite{BacteriaDetect}. On the account of the results obtained, we demonstrate designs that achieve accuracy comparable to the state-of-the-art~\cite{badrinarayanan2017segnet,kendall2015bayesian,jegou2017one,gal2017concrete,mehta2018espnet,yu2018bisenet,zhao2018icnet,nekrasov2020template}, but requiring only a fraction of the parameters or operations. Code for the implementation is at: \url{https://git.io/JmhTo}.

%% file: related_work.tex
\section{Related Work}\label{sec:related_work}

CNN-based architectures for image segmentation comprise of an encoder-decoder network, which first encodes the input into features and an upsampler that then recovers the output from the features as the decoder~\cite{long2014fully,badrinarayanan2017segnet}. The decoder is usually hierarchically opposite to the encoder, although both consist of multiple levels of computationally-expensive convolutions. Based on this encoder-decoder structure, the input is thereby refined to obtain the segmentation mask.

Long \textit{et al.}~\cite{long2014fully} first proposed the idea of Fully Convolutional Network (FCN) for this task, which outputs a segmentation mask in any given spatial dimensionality. Further improvements were achieved using bipolar interpolation and skip connections~\cite{he2016deep}. However, FCN is limited to few-pixel local information and therefore prone to lose global semantic context. SegNet~\cite{badrinarayanan2017segnet} was the first CNN trained end-to-end for segmentation. The novelty of the architecture was in eliminating the need for learning to upsample using fixed bilinear interpolation for resolution recovery. Ronneberger \textit{et al.}~\cite{ronneberger2015u} introduced a contracting and expansive pathway to better capture context and improve localisation precision, forming the characteristic "U"-shaped network. 

Atrous convolutions~\cite{chen2018encoder,zhao2017pyramid} have also been key to recent advancements, as they allow increasing the receptive field without changing the feature map resolution. Multiple such convolutional layers, that can accept the input in parallel, allow us to better account for multi-scale contextual information across images. This is termed Atrous Spatial Pyramid Pooling (ASPP)~\cite{zhao2017pyramid}. 

The downsampling of input images in deep classification networks can be hardware-inefficient, and several works have addressed this in the context of embedded vision applications. MobileNets~\cite{howard2017mobilenets} introduced the idea of factorising the standard convolution into depth-wise/kernel-wise separable convolutions, formed of a depth-wise convolution layer that filters the input and a $1\times1$ convolution that combines these to create new features. In~\cite{romera2017erfnet}, the authors employed kernel-wise separable convolutions to construct a compact model with the objective of enabling efficient real-time semantic segmentation. ESPNet~\cite{mehta2018espnet} used a hierarchical pyramid of dilated and $1\times 1$ convolutions to reduce the architecture size. Nekrasov \textit{et al.}~\cite{nekrasov2020template} developed an automatic way to find extremely light-weight architectures for image segmentation.

Bayesian neural networks~\cite{gal2017concrete,gal2015bayesian,mcallister2017concrete,liang2018bayesian,kendall2015bayesian} assign a probability distribution on the network weights instead of point estimates, to provide uncertainty measurements in the predictions. Employing this Bayesian mathematical grounding for CNNs enables us to obtain both the mask and uncertainty associated with it in the context of image segmentation. To the best of our knowledge, there are only two works focusing on 2D Bayesian CNNs in image segmentation for robust uncertainty quantification. Both of these approaches use Monte Carlo Dropout (MCD)~\cite{gal2016dropout}, in which Gal and Ghahramani cast dropout~\cite{srivastava2014dropout} training in a NN as Bayesian inference without the need for additional parametrisation. In~\cite{kendall2015bayesian} the authors searched and utilised dropout positioning in a SegNet~\cite{badrinarayanan2017segnet}. In~\cite{gal2017concrete} the authors learned the dropout rates with respect to a DenseNet-like architecture~\cite{jegou2017one}. 

In comparison to the related work, our work repurposes existing approaches ~\cite{he2016deep,jegou2017one,chen2018encoder,ronneberger2015u,howard2017mobilenets,kendall2015bayesian} to construct hardware-efficient 2D segmentation networks by decreasing the number of parameters and multiply-add-accumulate (MAC) operations while also providing improvements in accuracy. Furthermore, unlike previous hardware-efficient works, we use MCD for uncertainty quantification. 

%% file: method.tex
\section{ComBiNet}\label{sec:combinet}

The 2D network architecture of ComBiNet is presented in Figure~\ref{fig:architecture}, which is based on a "U"-net-like architecture~\cite{jegou2017one,ronneberger2015u} that divides itself into upsampling and downsampling paths as briefly described in Section~\ref{sec:related_work}. Skip connections connecting the paths preserve sharp edges by reducing the coarseness of masks and as a result contextual information from the input images can be preserved. The general building unit of the network is a Repeat block. A Dense block at the bottom of the network is used to capture global image features in addition to the optional ASPP blocks that are placed with skip connections. The input is processed using a $3\times 3$ 2D convolution Pre-processing block, while the output is processed through a $1 \times 1$ 2D convolution Post-processing block.

\subsection{Repeat Block}\label{sec:combinet_repeat_block}
Repeat blocks have the dual purpose of extracting features, through the Dense block, and extracting contextual information, through an optional ASPP block. Each block spatially downsamples the input by a factor of $2$ and later upsamples it back to the block's input resolution. 

The Repeat block is reusable, such that multiple blocks can be appended to one another to extract contextually richer features. The output of the Downsampling block is the input into the next Repeat block. This means the features and the input are processed at different spatial sizes. It is important to highlight there is a connection between the input of the Repeat block and the output of its encoding Dense block, prior to Downsampling. The input is concatenated to the output of the block, without being processed through the feature extracting Dense block, to enable propagation of local and global contextual information. 

\subsubsection{Basic Building Blocks}\label{sec:combinet_repeat_block_basic_building}
The Dense block is inspired by~\cite{huang2017densely,jegou2017one} and shown expanded in Figure~\ref{fig:architecture} on the right. It is a gradual concatenation of previous features allowing for feature-map changes processed through a Basic Layer (BL). A BL accepts inputs from all previous layers in a Dense block. The output channel number of the BL is restricted to a growth rate of $k$, which is constant for all BLs in the network, to avoid exponential increase in the channels propagated. More intuitively, it regulates the amount of new information each layer can contribute to the global state. For similar reasons, the output of the Dense block does not automatically include the original input, unless considering the downsampling path. The Dense block can have an arbitrary number of BLs and their counts are increased towards smaller spatial input size.
Efficient gradient and feature propagation is ensured by concatenations between all previous stages and the current stage. Details of the serially connected individual operations of BL, Downsampling and Upsampling are given below.

\begin{itemize}
    \item[] \textbf{Basic Layer}: Batch normalisation; ReLU; $3\times3$ Completely separable convolution; Dropout 
    \item[] \textbf{Downsample}: Batch normalisation; ReLU; $1\times1$ Convolution; Dropout; $2\times2$ Max-pooling with stride 1; $2\times2$ Blur with stride 2
    \item[] \textbf{Upsample}: Bilinear interpolation; $1\times1$ Convolution
\end{itemize}

The BL first performs batch normalisation (BN) which pre-processes the inputs coming from the different BLs. This operation is followed by ReLU and a $3\times3$ completely separable convolution for feature extraction. It consists of serially connected $1\times3; 3 \times 1$ convolutions with the output channel size same as the input, while being channel-wise separated, followed by a reshaping pointwise $1\times1$ convolution. We use completely separable convolutions for their parameter and MAC operation count efficiency. In particular, when paired with an appropriate $k$, BL can be an extremely compact feature extractor.  
The convolution is followed by a 2D dropout to provide regularisation and perform Bayesian inference~\cite{gal2015bayesian}. 

The Downsampling extracts coarse semantic features. The combined operations include BN, ReLU, $1\times 1$ convolution, dropout and $2\times 2$ max-pooling with stride 1 and $2\times 2$ blurring with stride 2. We used additional blurring with max-pooling to preserve shift-invariance of convolutions~\cite{zhang2019making}.

The Upsampling uses the parameter-less bilinear interpolation to save computational and memory resources. Furthermore, it also preserves shift invariance of objects in the input images and avoids aliasing~\cite{zhang2019making}. We add a $1\times 1$ 2D convolution to the output of the interpolation to refine the upsampled features.

\subsubsection{Atrous Spatial Pyramid Pooling (ASPP)}\label{sec:combinet_aspp}
ASPP~\cite{zhao2017pyramid,chen2018encoder}, as briefly introduced in Section~\ref{sec:related_work}, has been successfully used in various segmentation models to capture contextual information. It consists of atrous (dilated) convolutions which enables the preservation of shift-invariance while at the same time increasing the receptive field and enhancing the robustness to augmentations~\cite{zhang2019making}. Specifically, it is composed of $4$ convolutions interleaved with BN and ReLU to extract information over a wide spatial range though setting wider dilation rates in convolutions. Global average pooling and $1\times1$ convolution are used for global feature aggregation at the given scale. 
Each part accepts inputs from all channels, downscales them such that the output is only $32$ channels. These are concatenated with all other in the channel-dimension and refined to the output channel dimension by $1\times 1$ convolution. Finally, we regularise by applying dropout. In our work we use ASPP blocks in all Repeat blocks, except the first and the last one. We also changed the original ordering of the dilated convolutions to place BN first, instead of the convolution, for better regularisation. We kept the partial channel numbers to $32$ to limit computation.

\subsection{Bayesian Inference}\label{sec:combinet_bayesian_inference}

MCD~\cite{gal2016dropout,gal2015bayesian} provides a scalable way to learn a predictive distribution, by applying dropout~\cite{srivastava2014dropout} to the output of convolutions at both training and test time. This leads to Bayesian inference over the network’s weights. The sampled distribution provided by the dropout is used to sample models from the learnt variational posterior distribution. Although this can be achieved without additional parameters, it requires sampling and repeating $S$ feedforward steps through the network with the same input. The $S$ repeated steps linearly increase the compute demand, such that the runtime computational or memory complexity scales with $\mathcal{O}(S)$, and hence it is of further importance that the network is hardware efficient both in terms of memory consumption as well as the number of operations for the individual runs. A pixel-wise entropy can be derived, based on the repeated runs, that quantifies uncertainty as $\mathbb{H}(\hat{\boldsymbol{y}})=-\sum_c^C \hat{y}_c\log(\hat{y}_c)$. The $\hat{\boldsymbol{y}}\in\mathbb{R}^C$ is the pixel-wise mean of the softmax outputs across the $S$ runs with respect to $C$ output classes. The dropout rate presents a trade-off between data fit and uncertainty estimation. For convenience of hardware implementation, we use a dropout rate of 0.05 across the entire network for all experiments. 

%% file: experiments.tex
\section{Experiments}\label{sec:experiments}
This Section first discusses our experimentation settings and then presents an assessment of the results on the CamVid and bacteria datasets. We did not perform pre-training on additional image data or post-training fine-tuning. We introduce three ComBiNet models: ComBiNet-S, ComBiNet-M, ComBiNet-L: small, medium or large depending on the MAC count or the number of parameters, with the aim to trade-off computational complexity, accuracy and uncertainty quantification capabilities. 
We evaluated uncertainty through the mean per-pixel entropy of networks trained on CamVid or bacteria with respect to a random subset of 250 PascalVOC images~\cite{everingham2010pascal}. We initalised the weights of all ComBiNets with respect to the He-Uniform initialisation~\cite{he2016deep}. To train, we used Adam for 800 epochs with an initial learning rate of $0.001$ and an exponential decrease with a factor $0.996$. We trained ComBiNets with respect to a batch size of $2$ and with BN applied to each batch individually, as we found it essential to not use train-time statistics during evaluation. We set $S=30$ for the quantitative and qualitative software evaluation. For quantitative evaluation we measured the standard per-pixel mean intersection over union (mIoU), entropy, MACs and number of trainable parameters. The number of MACs was calculated with respect to $224 \times 224 \times 3$ input size and $S=1$. We repeated each experiment 3 times from which we report mean and $\pm$ a single standard deviation in following Tables.
\begin{figure}
     \centering
     \includegraphics[width=\linewidth]{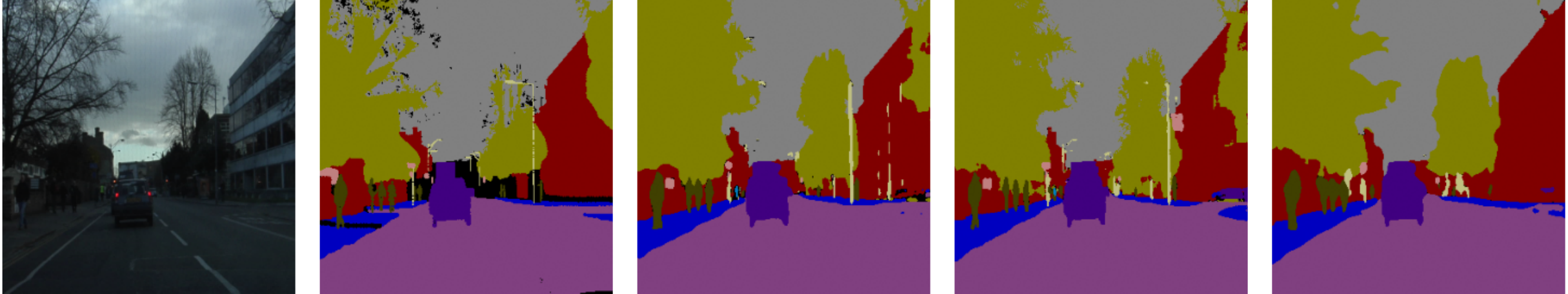}
     \caption{Qualitative evaluation. (from left) The first column depicts the input image, the second column is the ground-truth segmentation mask, the remaining columns are with respect to predictions of ComBiNet-L, DenseNet-103 + CD and DeepLab-v3+-ResNet50.}
     \label{fig:qualitative_camvid}
\end{figure}

\subsection{CamVid}\label{sec:experiments_camvid}
The CamVid road scenes dataset~\cite{BrostowSFC:ECCV08} originates from fully segmented videos from the perspective of a driving car. It consists of 367 frames for training, 101 frames for validation and 233 frames for testing of RGB images with a $480\times 360$ input resolution. There are 11 manually labelled classes that include roads, cars, signs etc. and a background that is usually ignored during training and evaluation. To augment the dataset we carried out channel-wise normalisation and the following randomly: re-scale inputs between a factor of $0.5$ to $2.0$; change aspect ratios between $3/4$ to $4/3$;  crop with a square size of $360$; horizontal flips; and random colour changes with respect to contrast, saturation and hue for training. We used the combo loss function~\cite{taghanaki2019combo}, and weighted it proportionally to class-pixels in the images as CamVid is unbalanced. A weight decay of $1e^{-3}$ was applied.

We summarise the performance of the different ComBiNets in Table~\ref{tab:camvid}, comparing to the other state-of-the-art 2D segmentation networks that include those focused on hardware efficiency with respect to their number of parameters and those considering Bayesian inference. The results show all ComBiNets obtained competitive results on mIoU with significantly fewer parameters and MACs. One result that stands out is~\cite{zhu2019improving} which used video, fine-tuning and an overparametrised architecture. ComBiNet-L is the most accurate of ComBiNets with approximately $3\times$ fewer parameters and MACs than its current equivalent with $S=1$. ComBiNet-S is the most hardware efficient with $42\times$ fewer parameters and $7\times$ fewer MACs than the Bayesian SegNet when $S=1$, while achieving an accuracy that is still close to the related works. We also compared the entropy pixel-wise, in which ComBiNets are marginally better in comparison to~\cite{kendall2015bayesian,gal2017concrete}. In Figures~\ref{fig:qualitative_camvid} and~\ref{fig:qualitative} we demonstrate the qualitative results. In general, the model is more uncertain in the objects that are more distant, occluded or surrounded by the background class (black), which was ignored during training and evaluation. The results of the segmentation showed that the most problematic classes were fence and sign/symbol, whilst roads and the sky were most accurately distinguished. Figure~\ref{fig:qualitative_camvid} demonstrates on one sample that the model is accurate also in comparison to the related work consisting of a non-Bayesian or a Bayesian model.

\begin{table}
  \caption{Comparison with respect to other networks on the CamVid test dataset, $\dagger$ notes training and testing with respect to $960\times720$ images instead of $480\times360$, $\ddagger$ denotes Bayesian approaches. Arrows denote desired trends. - denotes not reported. $*$ were replicated in this work and not officially reported.}
  \label{tab:camvid}
  \scalebox{.97}{
  \begin{tabular}{|l|l|l|l|l|}
    \hline
     \textbf{Method}&\textbf{mIoU} [\%] $\uparrow$& \textbf{Params} [M] $\downarrow$ & \textbf{MACs} [G] $\downarrow$ & \textbf{Entropy} [nats] $\uparrow$\\
    \hline
    SegNet~\cite{badrinarayanan2017segnet} & 55.6& 29.7 & - & -\\
    Bayesian SegNet$^\ddagger$~\cite{kendall2015bayesian}& 63.1& 29.7 & 30.8$^*$ & 0.68$^*$ \\
    DenseNet-103~\cite{jegou2017one}& 66.9& 9.4 & 24.9$^*$& -\\
    DenseNet-103 + CD$^\ddagger$~\cite{gal2017concrete}& 67.4& 9.4 & 24.9$^*$ & 0.47$^*$ \\
    ESPNet~\cite{mehta2018espnet} & 55.6& 0.36 & -& -\\
    BiSeNet$^\dagger$~\cite{yu2018bisenet} & 65.6& 5.8 & -& -\\
    ICNet$^\dagger$~\cite{zhao2018icnet} & 67.1& 6.7 & -& -\\
    Compact Nets~\cite{nekrasov2020template} & 63.9& \textbf{0.28} & -& -\\
    DeepLab-v3+-ResNet50~\cite{chen2018encoder} & 57.6$^*$& 16.6$^*$& 13.2$^*$& -\\
    Video-WideResNet38$^\dagger$~\cite{zhu2019improving} & \textbf{79.8} & 137.1& - & - \\
    \hline
    ComBiNet-S$^\ddagger$ & 66.1$\pm$0.3 & 0.7 & \textbf{4.2}  & \textbf{0.69$\pm$0.02}\\
    ComBiNet-M$^\ddagger$ & 66.9$\pm$0.2 & 1.3 & 7.9 & 0.68$\pm$0.01\\
    ComBiNet-L$^\ddagger$ & 67.9$\pm$0.1 & 2.3 & 9.4 & 0.65$\pm$0.02\\
  \hline
\end{tabular}}
\end{table}

\subsection{Bacteria}\label{sec:experiments_bacteria}
The bacteria dataset~\cite{BacteriaDetect} comprises of 366 darkfield microscopy images with manually annotated masks for segmentation. The task is to detect bacteria of the phylum Spirochaetes in blood. This therefore leads to a problem of segmenting two classes corresponding to the bacteria and red blood cells - Spirochaetes and Erythrocytes respectively. This is a challenging task due to both the nature of the problem, a heavily unbalanced dataset, and the collection methodology which results in considerable noisy RGB input images of varying sizes from $1000\times 1000$ to $300\times 300$ pixels. We randomly split the dataset into sizes 219, 73, 74 images for training, validation and test respectively. We then apply the same augmentations as those mentioned in Section~\ref{sec:experiments_camvid} for the CamVid dataset, extended further with vertical flips. We train with respect to the Combo loss function and added a log-dice coefficient. Weight decay was set to $1e^{-4}$.
\begin{table}
  \caption{Comparison with respect to other networks on the bacteria test dataset, $\ddagger$ denotes Bayesian approaches. Arrows denote desired trends. $*$ were replicated in this work and not officially reported.}
  \label{tab:bacteria}
  \scalebox{.95}{
  \begin{tabular}{|l|l|l|l|l|}
    \hline
     \textbf{Method}&\textbf{mIoU} [\%] $\uparrow$& \textbf{Params} [M] $\downarrow$ & \textbf{MACs} [G] $\downarrow$ & \textbf{Entropy} [nats] $\uparrow$\\
    \hline
    Bayesian SegNet$^{\ddagger, *}$~\cite{kendall2015bayesian}& 76.1& 29.7 & 30.8 & 0.19 \\
    DenseNet-103 + CD$^{\ddagger, *}$~\cite{gal2017concrete}& 75.8& 9.4 & 24.9 & \textbf{0.32} \\
    U-Net$^{*}$~\cite{ronneberger2015u} & 71.4 & 31.0 & 41.9& -\\
    DeepLab-v3+-ResNet50$^{*}$~\cite{chen2018encoder} & 80.4 & 16.6 & 13.2& -\\
    \hline
    ComBiNet-S$^\ddagger$ & 82.3$\pm$0.4 & \textbf{0.7} & \textbf{4.2}  & 0.18$\pm$0.02\\
    ComBiNet-M$^\ddagger$ & \textbf{83.0$\pm$0.4} & 1.3 & 7.9 & 0.16$\pm$0.01\\
    ComBiNet-L$^\ddagger$ & 82.3$\pm$0.2 & 2.3 & 9.4 & 0.16$\pm$0.02\\
  \hline
\end{tabular}}
\end{table}
\begin{figure}
     \centering
     \includegraphics[width=\linewidth]{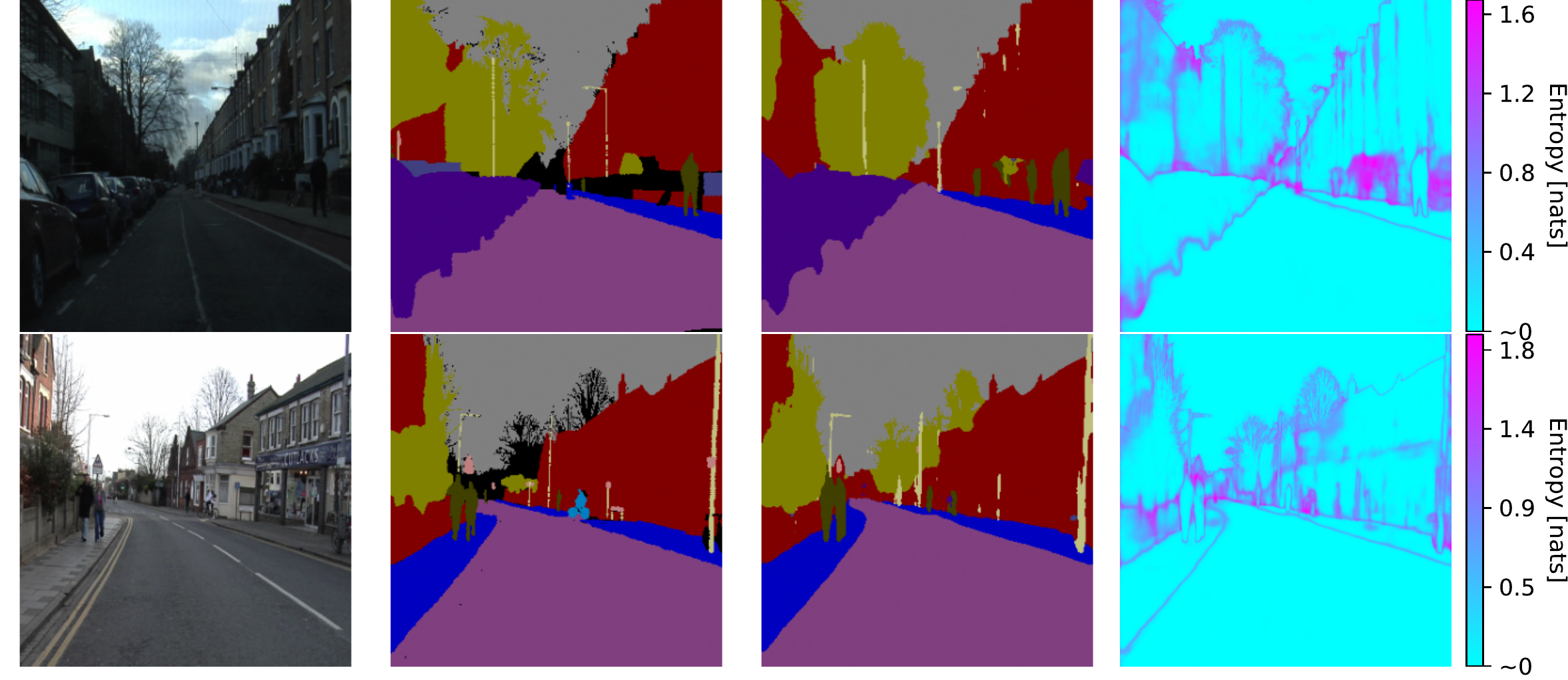}
     \newline
     \includegraphics[width=\linewidth]{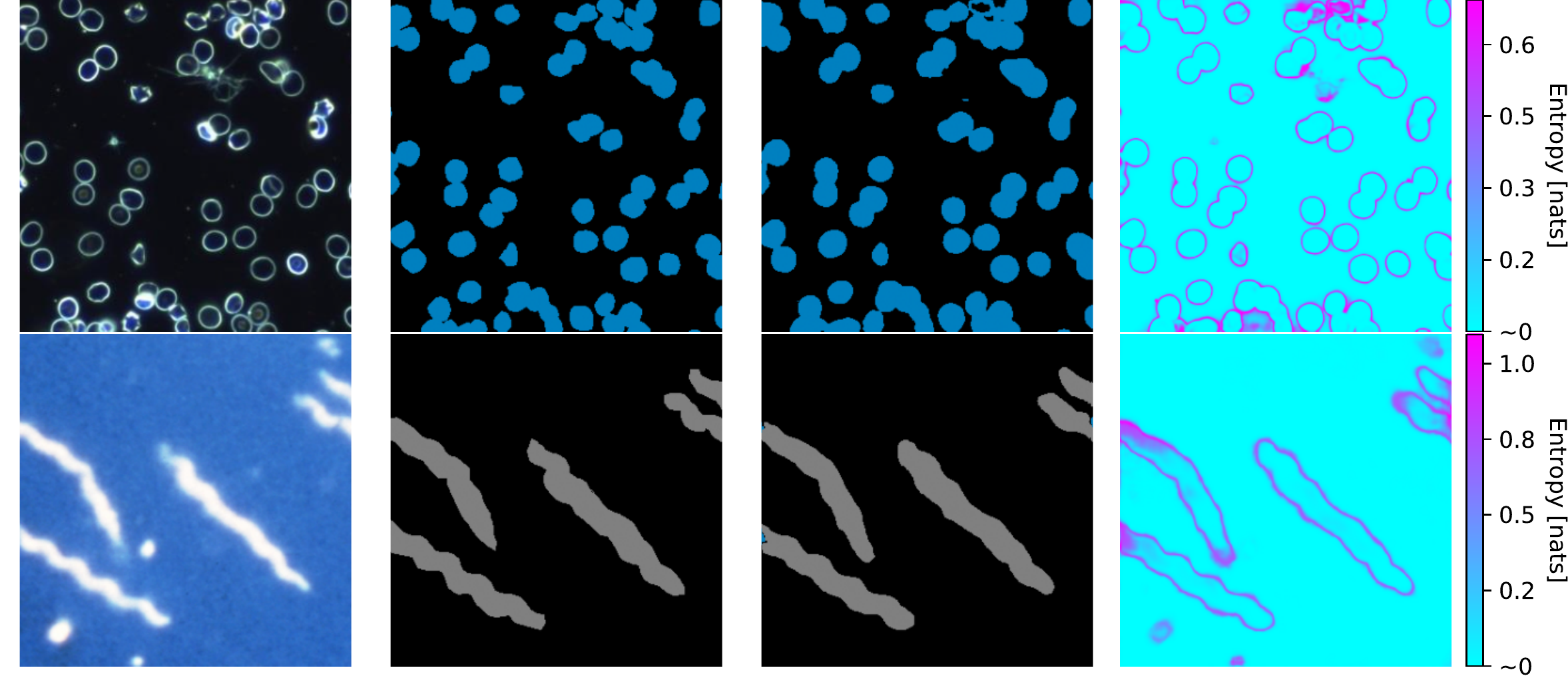}
     \caption{Qualitative evaluation. (from left) The first column depicts the input image, the second column are the ground-truth segmentation masks, the third column are the predictions and the fourth column are the per-pixel uncertainties measured through the predictive entropy of ComBiNet-L and ComBiNet-M. The two top rows are with respect to CamVid models and the bottom two rows are with respect to bacteria models.}
     \label{fig:qualitative}
\end{figure}

\begin{figure}[t]
     \centering
     \includegraphics[width=\linewidth]{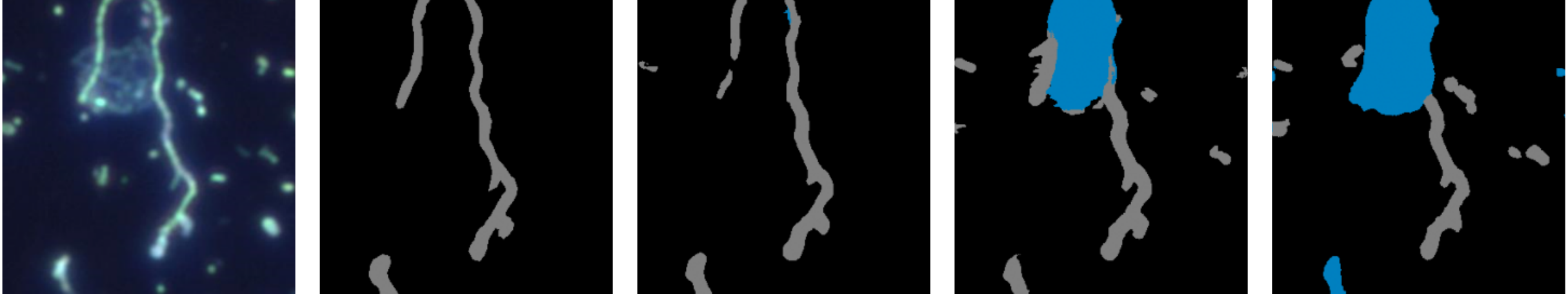}
     \caption{Qualitative evaluation. (from left) The first column depicts the input image, the second column is the ground-truth segmentation mask, the remaining columns are with respect to predictions of ComBiNet-M, DenseNet-103 + CD and Bayesian SegNet.}
     \label{fig:qualitative_bacteria}
\end{figure}

\begin{figure}[t]
     \centering
     \includegraphics[width=\linewidth]{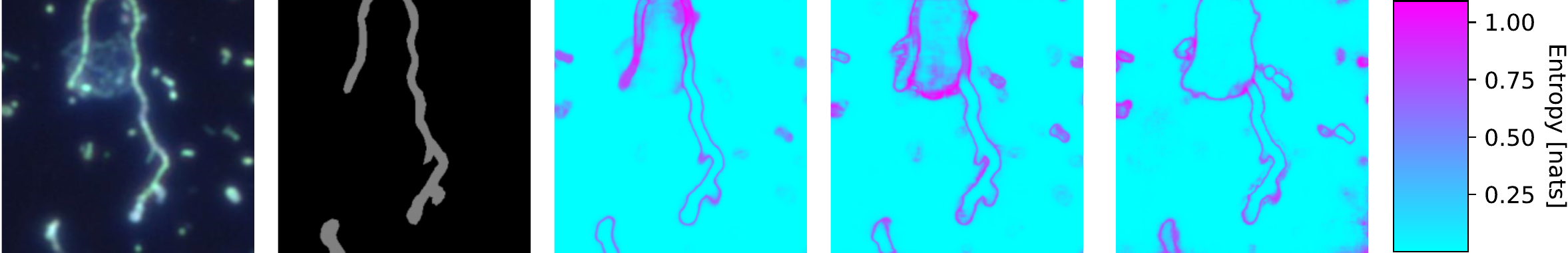}
     \caption{Qualitative evaluation of predictive uncertainty. (from left) The first column depicts the input image, the second column is the ground-truth segmentation mask and the remaining columns are with respect to predictions of ComBiNet-M, DenseNet-103 + CD and Bayesian SegNet.}
     \label{fig:qualitative_bacteria_uncertainty}
\end{figure}

Table~\ref{tab:bacteria} shows that all ComBiNets obtain better accuracy with significantly fewer parameters and MACs. ComBiNet-S is the most hardware efficient with $13\times$ fewer parameters and $6\times$ fewer MACs than DenseNet when $S=1$. We note that ComBiNet-87 achieves a worse accuracy than ComBiNet-M in our experiments with this dataset, showing that a bigger network is not always the best. 
All ComBiNets infer that all unrecognisable objects should be classified as a background resulting in smaller entropy than the related work. The qualitative evaluation of Figures~\ref{fig:qualitative} and~\ref{fig:qualitative_bacteria} demonstrates the ability of the architecture to segment noisy images, while comparing it to DenseNet with Concrete Dropout (CD)~\cite{gal2017concrete} and Bayesian SegNet. We further depict the corresponding predictive uncertainty of this sample in Figure~\ref{fig:qualitative_bacteria_uncertainty}, which helps us understand the portions of the image where the architecture was less certain in its given predictions. It can be seen that the network is uncertain about suspicious bacteria bodies, which can further help practitioners to better understand their samples. 

\subsection{Discussion}
With respect to the qualitative results in Figure~\ref{fig:qualitative} along with the quantified uncertainty measured by per-pixel entropy we observe that, due to the skip connections and gradual downsampling and upsampling, the model retains sharp edges and detail in the predictions. Additionally, through feature reuse and detailed construction of efficient modules, e.g. BL, the model was able to provide favourable performance, despite smaller parameter or operation counts.

The main bottleneck of this work lies in its use of MCD for Bayesian inference, as it requires multiple feedforward runs, but no extra network weights, to obtain an uncertainty estimate in the output mask. These runs multiply the MAC cost $\times S$ and hence $S$ represents a trade-off between hardware demand and quality of approximation of the predictive distribution. For this reason lowering MACs at the individual feedforward pass level was the focus of this work. Additionally, in hardware it is possible to simply parallelise these runs~\cite{myojin}. Lastly, if uncertainty estimation is not needed, the presented networks can still guarantee high accuracy with respect to weight averaging, disabling dropout and setting $S=1$, which was relatively lower by approximately one standard deviation as shown in the Tables~\ref{tab:camvid} and~\ref{tab:bacteria} for CamVid or bacteria respectively.

%% file: conclusion.tex
\section{Conclusion}\label{sec:conclusion}

We propose a compact 2D Bayesian architecture, ComBiNet, that re-purposes hardware efficient operations for the task of image segmentation. We demonstrated that good accuracy along with predictive uncertainties can be achieved with significantly fewer parameters and MACs, lowering hardware resources and computational costs. We show that ComBiNet performs well with an imbalanced dataset, as well as the established CamVid dataset, showing higher uncertainty in misclassified sections. Furthermore, it was not necessary to perform any pre-training or post-training fine-tuning to reach the observed accuracy. For the future, we would like to measure and optimise the architectures with respect to other hardware performance metrics such as power consumption or structured instance-wise uncertainty estimation instead of pixel-wise.